\documentclass[conference, 10pt]{IEEEtran} 
\usepackage[compress]{cite}

\ifCLASSINFOpdf
\else
\fi
\usepackage[cmex10]{amsmath}
\usepackage[caption=false,font=footnotesize]{subfig}
\usepackage{url}

\usepackage{amssymb, amscd}
\usepackage{mathtools}

\usepackage{pgf,tikz}
\usepackage{pgfplots}

\usetikzlibrary{external}
\usepackage{hyperref}
\hypersetup{
    bookmarks=true,         
    unicode=false,          
    pdftoolbar=true,        
    pdfmenubar=true,        
    pdffitwindow=false,     
    pdfstartview={FitH},    
    pdfauthor={Michael Tschannen},     
    pdfsubject={Subject},   
    pdfcreator={Michael Tschannen},   
    pdfproducer={Producer}, 
    pdfnewwindow=true,      
    colorlinks=false,       
    linkcolor=red,          
    citecolor=green,        
    filecolor=magenta,      
    urlcolor=cyan           
}

\newcounter{AlgCount}

 \newcommand{\mc}[0]{\mathcal }
 \newcommand{\mb}[0]{\mathbb }

\usepackage{framed}

\newcommand\norm[2][\Tnorm]{\ensuremath{{\left\Vert #2 \right\Vert}_{#1}}}

\newcommand\Tinnerprod{}
\newcommand{\innerprod}[3][\Tinnerprod]{\ifthenelse{\equal{#1}{}}{\ensuremath{\left<#2,#3\right>}}{\ensuremath{\left<#2,#3\right>_{#1}}}}

\newcommand\vect[1]{\mathbf #1}

\definecolor{DarkBlue}{rgb}{0.1,0.1,0.5}
\definecolor{BrickRed}{RGB}{203,65,84}

\newcommand\PR[1]{\ensuremath{ {\mathrm{P}}\!\left[#1\right]}}

\newcommand\Tex{}
\newcommand\EX[2][\Tex]{
\ifthenelse{\equal{#1}{}}{{\mathbb E}\left[#2\right]}{\ensuremath{{\mathbb E}_{#1}\left[ #2\right]}}}

\newcommand\Var[2][\Tex]{
\ifthenelse{\equal{#1}{}}{{\mathrm{Var} }[#2]}{\ensuremath{\mathrm{Var}_{#1}\left[ #2\right]}}}

\newcommand\ignore[1]{}

\newcommand\defeq{\coloneqq}
\newcommand\eqdef{\eqqcolon}

\newcommand{\reals}{\mathbb R} 
 
\newcommand{\transp}[1]{{#1}^T} 

\renewcommand{\d}{d} 

\renewcommand{\d}{d} 
\newcommand{\X}{\mathcal X}

\newcommand{\va}{\vect{a}}  
\newcommand{\vb}{\vect{b}}

\newcommand{\vv}{\vect{v}}  

\newcommand{\vx}{\vect{x}}  
\newcommand{\vy}{\vect{y}}  
\newcommand{\vz}{\vect{z}}

\newcommand{\mA}{\vect{A}}  
\newcommand{\mB}{\vect{B}} 
\newcommand{\mC}{\vect{C}}

\newcommand{\mG}{\vect{G}}

\newcommand{\mI}{\vect{I}}

\newcommand{\mQ}{\vect{Q}}
\newcommand{\mR}{\vect{R}}

\newcommand{\mZ}{\vect{Z}}

\renewcommand{\l}{\ell}
\newcommand{\powpar}[2]{{#1}^{(#2)}}
\newcommand{\Xl}{\powpar{X}{\l}}
\newcommand{\Xk}{\powpar{X}{k}}

\newcommand{\Xlt}{\powpar{\tilde X}{\l}}
\newcommand{\xlh}{\powpar{\hat x}{\l}}
\newcommand{\xkh}{\powpar{\hat x}{k}}

\newcommand{\sk}{\powpar{s}{k}}
\renewcommand{\sl}{\powpar{s}{\l}}
\newcommand{\slt}{\powpar{\tilde s}{\l}}
\newcommand{\slh}{\powpar{\hat s}{\l}}
\newcommand{\skh}{\powpar{\hat s}{k}}
\newcommand{\ek}{\powpar{e}{k}}
\newcommand{\el}{\powpar{e}{\l}}
\newcommand{\rl}{\powpar{r}{\l}}
\newcommand{\rlt}{\powpar{\tilde r}{\l}}
\newcommand{\xh}{\hat x}
\newcommand{\rh}{\hat r}
\newcommand{\rlh}{\powpar{\hat r}{\l}}
\newcommand{\sh}{\hat s}

\newcommand{\wk}{\powpar{w}{k}}
\newcommand{\wl}{\powpar{w}{\l}}
\newcommand{\mul}{\powpar{\mu}{\l}}

\newcommand{\al}{\powpar{\alpha}{\l}}

\newcommand{\cX}{\mc X}

\renewcommand{\d}{\mathrm{d}}
\renewcommand{\i}{\mathrm{i}}
\newcommand{\Wl}{\powpar{W}{\l}}

\newcommand{\abs}[1]{\left\vert #1 \right\vert}
\newcommand{\abss}[1]{\vert #1 \vert}

\newcommand{\E}[1]{\mb E \! \left[ #1 \right]}

\AtBeginDocument{
  \addtolength\abovedisplayskip{-2.5pt}
  \addtolength\belowdisplayskip{-2.5pt}
}

\usepackage{amsthm}
\makeatletter
\def\thm@space@setup{%
  \thm@preskip=4pt \thm@postskip=\thm@preskip
}

\newtheorem{thm}{Theorem}

\newtheorem{definition}{Definition}

\begin{document}
%

\title{Nonparametric Nearest Neighbor Random Process Clustering}

\author{
	\IEEEauthorblockN{Michael Tschannen and Helmut B{\"o}lcskei} \\[-0.25cm]
	\IEEEauthorblockA{Dept. IT \& EE, ETH Zurich, Switzerland \\ Email: \{michaelt, boelcskei\}@nari.ee.ethz.ch}
}


\maketitle

\begin{abstract}

We consider the problem of clustering noisy finite-length observations of stationary ergodic random processes according to their nonparametric generative models without prior knowledge of the model statistics and the number of generative models. Two algorithms, both using the $L^1$-distance between estimated power spectral densities (PSDs) as a measure of dissimilarity, are analyzed.
The first algorithm, termed nearest neighbor process clustering (NNPC), to the best of our knowledge, is new and relies on partitioning the nearest neighbor graph of the observations via spectral clustering. The second algorithm, simply referred to as $k$-means (KM), consists of a single $k$-means iteration with farthest point initialization and was considered before in the literature, albeit with a different measure of dissimilarity and with asymptotic performance results only.
We show that both NNPC and KM succeed with high probability under noise and even when the generative process PSDs overlap significantly, all provided that the observation length is sufficiently large. Our results quantify the tradeoff between the overlap of the generative process PSDs, the noise variance, and the observation length. Finally, we present numerical performance results for synthetic and real data.
\end{abstract}


%
\IEEEpeerreviewmaketitle

\section{Introduction}

Consider a set of $N$ noisy length-$M$ observations of stationary ergodic random processes stemming from $L < N$ (typically $L \ll N$) different generative models. We want to cluster these observations according to their generative models without prior knowledge of the model statistics and the number of generative models $L$.
This problem arises in many domains of science and engineering where large amounts of data have to be divided into meaningful subsets in an unsupervised fashion, e.g., for efficient processing or storage. Examples include clustering of audio and video sequences \cite{wang2000multimedia}, electrocardiography (ECG) recordings \cite{kalpakis2001distance}, and industrial production indices \cite{corduas2008time}.

Many existing random process clustering methods quantify the dissimilarity between observations using the Euclidean distance between either estimated model parameters \cite{corduas2008time}, or estimated cepstral coefficients \cite{kalpakis2001distance, boets2005clustering}, or normalized periodograms \cite{caiado2006periodogram}. Other methods rely on divergences (e.g., Kullback-Leibler divergence) between normalized periodograms \cite{kakizawa1998discrimination, vilar2004discriminant} or use the distributional distance between processes \cite{ryabko2010clustering, khaleghi2012online}. In all cases the resulting distances are then fed to a standard clustering algorithm such as $k$-means or hierarchical clustering. Another line of work employs a Bayesian framework to infer the cluster assignments, e.g., according to a maximum a posteriori criterion \cite{xiong2004time}. While many of these approaches have proven effective in practice, corresponding analytical performance results are scarce. Moreover, existing analytical results are mostly concerned with the asymptotic regime where the observation length tends to infinity (see, e.g., \cite{kakizawa1998discrimination, vilar2004discriminant, corduas2008time, ryabko2010clustering, khaleghi2012online}); the finite observation-length regime has attracted virtually no attention.

\paragraph*{Contributions}
We consider two process clustering algorithms that apply to nonparametric generative models and measure dissimilarity between observations through the $L^1$-distance between estimated power spectral densities (PSDs). The first algorithm, termed nearest neighbor process clustering (NNPC), is inspired by the subspace clustering algorithm described in \cite{heckel2014robust} and, to the best of our knowledge, has not been considered in the literature before. NNPC relies on partitioning the $q$-nearest neighbor graph of the observations via spectral clustering (the number of nearest neighbors $q$ is a parameter of NNPC). The second algorithm, which will simply be referred to as $k$-means (KM), consists of a single $k$-means iteration with farthest point initialization \cite{katsavounidis1994new} and was first proposed in \cite{ryabko2010clustering} with a different distance measure. The original formulation of KM was shown in \cite{ryabko2010clustering} to 
deliver the correct segmentation with probability tending to $1$ as the observation length $M \to \infty$.

Assuming real-valued zero-mean stationary ergodic Gaussian processes as generative models, we characterize the performance of NNPC and KM analytically for finite-length observations contaminated by independent additive white Gaussian noise. We find that both algorithms succeed with high probability even when the (true) PSDs exhibit significant overlap, all provided that the observation length is sufficiently large. Our analytical results quantify the tradeoff between observation length, noise variance, and distance between the (true) PSDs of the generative models. Finally, we evaluate the performance of the two algorithms on synthetic and on real data.

\paragraph*{Notation} 
We use lowercase boldface letters to denote vectors, uppercase boldface letters to designate matrices, and the superscript ${}^T$ stands for transposition. $v_i$ is the $i$th entry of the vector $\vv$. For the matrix $\mA$, $\mA_{i,j}$ designates the entry in row $i$ and column $j$, $\norm[2 \to 2]{\mA}$ its maximum singular value, and $\norm[F]{\mA} \defeq (\sum_{i,j} \abs{\mA_{i,j}}^2)^{1/2}$ its Frobenius norm. $\mI$ stands for the identity matrix. The $i$th element of a sequence $x$ is denoted by $x[i]$. For a positive integer $N$, $[N]$ designates the set $\{1,2,\dots,N\}$. $\E{X}$ is the expectation of the random variable $X$ and the notation $Y \sim X$ indicates that the random variable $Y$ has the same distribution as $X$. We say that a subgraph $H$ of a graph $G$ is connected if every pair of nodes in $H$ can be joined by a path with nodes exclusively in $H$. A connected subgraph $H$ of $G$ is called a connected component of $G$ if there are no edges between $H$ and the remaining nodes in $G$ \cite{luxburg2007tutorial}.

\section{Formal problem statement and algorithms} \label{sec:ProbAlgo}

We consider the following clustering problem: Given the unlabeled data set $\cX = \cX_1 \cup \dots \cup \cX_L$ of cardinality $N$, where $\X_\l = \{ \xlh_i \}_{i=1}^{n_\l}$ contains the contiguous noisy length-$M$ observations $\xlh_i$ of the real-valued stationary ergodic random process $\Xl[m], m \in \mb Z,$ corresponding to the $\l$th generative model, find the partition $\cX_1,\dots,\cX_L$. The statistics of the generative models and of the noise processes, as well as the number of generative models, are all assumed unknown.

Both clustering algorithms we consider are based on the following distance measure. Denoting the PSD of $\Xl$ by $\sl(f)$, $f \in [0,1)$, we define the distance between $\Xk$ and $\Xl$ by $d(\Xk, \Xl) \defeq \frac{1}{2} \int_0^1 \abss{\sk(f) - \sl(f)} \d f$. For processes $\Xl$ with unit power, i.e., $\int_0^1 \sl(f) \d f = 1$, $\l \in [L]$, the factor $1/2$ ensures that $d(\Xk,\Xl)$ takes on values in $[0, 1]$. $d(\Xk,\Xl)$ essentially quantifies the dissimilarity between the support sets of $\sk$ and $\sl$, i.e., it is close to $1$ when $\sk$ and $\sl$ are concentrated on disjoint frequency bands.

We now present the NNPC and the KM algorithms.
Recall that NNPC is inspired by the subspace clustering algorithm introduced in \cite{heckel2014robust}, and KM is obtained by replacing the distance measure in Algorithm 1 in \cite{ryabko2010clustering} by the distance measure $d$ defined above.
\\[-0.3cm]

{\it \refstepcounter{AlgCount} \label{alg:TSC}
{\bf The NNPC algorithm. } Given a set $\cX$ of $N$ length-$M$ observations, the number of generative models $L$ (the estimation of $L$ from $\cX$ is discussed below), and the parameter $q$, carry out the following steps: \\
{\bf Step 1:} For every $\xh_i \in \cX$, estimate the PSD $\sh_i(f)$ via the Blackman-Tukey (BT) estimator according to
\begin{align}
\sh_i(f) &\defeq \sum_{m=-M+1}^{M-1} g[m] \rh_i[m] e^{-\i 2 \pi f m}, \quad \text{where} \label{eq:BlackmanTukey} \\
\rh_i[m] &\defeq \frac{1}{M} \! \! \sum_{n = 1}^{M-\vert m \vert} \xh_i[n+m] \xh_i[n], \quad \vert m \vert \leq M-1, \nonumber
\end{align}
and $g[m]$, $m \in \mb Z$, is an even window function (i.e., $g[m] = g[-m]$) with $g[0] = 1$ and nonnegative bounded discrete-time Fourier transform $g(f)$, i.e., $0 \leq g(f) \leq A < \infty$, $f \in [0,1)$. Identify the set $\mc T_i \subset [N] \setminus \{ i \}$ of cardinality $q$ defined through
\begin{equation}
d(\xh_i,\xh_j) \leq d(\xh_i,\xh_p), \quad \text{for all} \; j \in \mc T_i \; \text{and all} \; p \notin \mc T_i. \nonumber
\end{equation}
{\bf Step 2:} Let $\vz_j \in \reals^N$ be the vector with $i$th entry $\exp(-2 \, d(\xh_i,\xh_j))$, if $j \in \mc T_i$, and $0$, if $j \notin \mc T_i$. \\
{\bf Step 3:} Construct the adjacency matrix $\mA$ according to $\mA = \mZ + \transp{\mZ}$, where $\mZ = [\vz_1 \, \dots \, \vz_N]$. \\
{\bf Step 4:} Apply normalized spectral clustering \cite{luxburg2007tutorial} to $(\mA,L)$.}
\\[-0.3cm]

{\it \refstepcounter{AlgCount} \label{alg:kmeans}
{\bf The KM algorithm \cite{ryabko2010clustering}. } Given a set $\cX$ of $N$ length-$M$ observations and the number of generative models $L$, carry out the following steps: \\
\setlength{\parindent}{1.7cm}
\setlength{\parskip}{0cm}
{\bf Step 1:} Initialize $c_1 \defeq 1$ and $\hat \cX_\l \defeq \{\}$ for all $\l \in [L]$. \\ 
{\bf Step 2:} For every $\xh_i \in \cX$, estimate the PSD $\sh_i(f)$ via the BT estimator \eqref{eq:BlackmanTukey}. \\
{\bf Step 3:} {\bf for $p = 2$ to $L$ do:}

$c_p \defeq \arg \max_{i \in [N]} \, \min_{\l \in [p-1]} d(\xh_i,\xh_{c_\l})$. \\ 
{\bf Step 4:} {\bf for $i = 1$ to $N$ do:}

$\l^\star \gets \arg \min_{\l \in [L]} \, d(\xh_i,\xh_{c_\l})$

$\hat \cX_{\l^\star} \gets \hat \cX_{\l^\star} \cup \{ \xh_i \}$
}
\\[-0.3cm]

Both algorithms are based on comparing distances between observations, and are therefore meaningful only if the observations are of comparable average power.

We henceforth denote the nearest neighbor graph with adjacency matrix $\mA$ obtained in Step 3 of NNPC by $G$. The parameter $q$ in the NNPC algorithm determines the number of edges in $G$. Choosing $q$ too small may result in the observations stemming from a given generative model forming multiple connected components in $G$ and hence not being assigned to the same cluster. This problem can be countered by increasing $q$, which, however, increases the chances of observations coming from different generative models being connected in $G$ and hence being misclustered. The issue of how to choose $q$ in practice is further discussed in the next section.

In the NNPC algorithm $L$ may be estimated using the \emph{eigengap heuristic} \cite{luxburg2007tutorial}, which relies on the fact that the number of zero eigenvalues of the normalized Laplacian of $G$ corresponds to the number of connected components of $G$. 

We finally note that the BT PSD estimates \eqref{eq:BlackmanTukey} can be computed efficiently via the FFT.

\section{Performance results} \label{sec:mainres}

For our analytical performance results, we assume that the $\xlh_i$, for given $\l$, are obtained as contiguous length-$M$ observations of $\Xlt[m] \defeq \Xl[m] + \Wl[m], m \in \mb Z$, where $\Xl$ is zero-mean stationary Gaussian with PSD $\sl(f)$, and $\Wl$ is a zero-mean white Gaussian noise process with variance $\sigma^2$ and independent of $\Xl$. Note that the noise statistics are identical across the the generative models.
The autocorrelation functions (ACFs) $\rl[m] \defeq \int_0^1 \sl(f) e^{\i 2 \pi f m} \d f$ are assumed absolutely summable, i.e., $\sum_{m = -\infty}^\infty \abss{\rl[m]} < \infty$, $\l \in [L]$, which implies ergodicity of the corresponding $\Xl$. Further, we assume that the PSDs are normalized according to $\int_0^1 \sl(f) \d f = 1$, $\l \in [L]$, and we define $B \defeq \max_{\l \in [L]} \sup_{f \in [0,1)} \sl(f)$. Our performance results depend on the maximum ACF moment $\mu_{\max} \defeq \max_{\l \in [L]} \mul$, where $\mul \defeq \sum_{m=-\infty}^\infty \abss{h[m]} \abss{\rl[m]}$ with
\begin{equation} \label{eq:weightedwindow}
h(m) \defeq \left\{ \begin{matrix*}[l] 1 - g(m)(1 - \vert m \vert /M), & \text{for} \; \vert m \vert < M \\
1, & \text{otherwise.} \end{matrix*} \right.
\end{equation}
For each $\l$, the $\xlh_i$ may either stem from independent realizations of $\Xlt$ or correspond to different (possibly overlapping) length-$M$ segments of a given realization of $\Xlt$. In the latter case the $\xlh_i$ will not be statistically independent in general.

Our main result for NNPC ensures the \emph{no false connections (NFC) property} defined as follows.
\begin{definition}[No False Connections Property] 
$G$ has no false connections if, for all $\l \in [L]$, nodes corresponding to $\cX_\l$ are connected to other nodes corresponding to $\cX_\l$ only. 
\end{definition}
Although the NFC property alone does not guarantee correct clustering, it was found to be a sensible performance measure for subspace clustering algorithms (see, e.g., \cite{soltanolkotabi2014robust, heckel2014robust}). To ensure correct clustering one would additionally have to ensure that the subgraphs of $G$ corresponding to the $\cX_\l$ are connected. Proving connectivity for NNPC appears to be difficult in the noisy finite observation-length regime.

\begin{thm} \label{thm:NNPC}
Let $q \leq \min_{\l \in[L]} (n_\l - 1)$ and let $\cX$ be generated according to the data model described above. Then, the \emph{clustering condition}
\begin{align} 
\min_{\substack{k, \l \in [L] \colon \\ k \neq \l}} &d(\Xk, \Xl) \nonumber \\[-0.6cm]
& \qquad > 8 A (B+\sigma^2)\sqrt{\frac{2 \log M}{M}} +2 \mu_{\max} \label{eq:ClusCond}
\end{align}
guarantees that $G$ has NFC with probability at least $1- 2N/M^2$.
\end{thm}

Our main result for KM comes in terms of a stronger performance guarantee, namely it ensures correct clustering. This is thanks to the fact that KM does not have a spectral clustering step and is hence much easier to analyze. On the other hand NNPC typically outperforms KM in practice.

\begin{thm} \label{thm:KM}
Let $\cX$ be generated according to the data model described above. Then, under the clustering condition \eqref{eq:ClusCond}, the partition $\hat \cX_1,\dots, \hat \cX_L$ of $\cX$ inferred by KM corresponds to the true partition $\cX_1,\dots,\cX_L$ with probability at least $1- 2N/M^2$.
\end{thm}

The proofs of Theorems \ref{thm:NNPC} and \ref{thm:KM} are given in the Appendix. Theorems \ref{thm:NNPC} and \ref{thm:KM} essentially state that NNPC and KM succeed even when the PSDs $\sl$ of the $\Xl$ overlap significantly and the observations are contaminated by strong noise, all this provided that the observation length $M$ is sufficiently large and the window $g$ smoothes the BT PSD estimates appropriately so that $\mu_{\max}$ is small. The clustering condition \eqref{eq:ClusCond} quantifies the tradeoff between the (maximum) amount of overlap of the $\sl$ (through $\min_{k, \l \in [L] \colon k \neq \l} d(\Xk,\Xl)$), the observation length $M$, and the noise variance $\sigma^2$. The first term on the RHS of \eqref{eq:ClusCond} vanishes as $M \to \infty$. For $\rl$ with small essential support relative to $M$, i.e., $\rl[m] \approx 0$ for $m \geq M_0$ with $M_0 \ll M$, choosing $g$ such that $g[m] \approx 1$ for $m \leq M_0$ yields $\mu_{\max} \ll 1$ (since $h[m] \approx 1-g[m] \approx 0$ for $m \leq M_0$). Hence, the clustering condition \eqref{eq:ClusCond} can indeed be satisfied if the $\rl$ decay rapidly enough. Note that the choice of $g$ affects the constant $A$. In particular, increasing the width of $g$ results in larger $A$.

To ensure that the probability of success of NNPC and KM is high, we need to take $M \gg \sqrt{N}$, i.e., the observation length should be large relative to the square root of the number of observations. 

We next address the choice of $q$ for NNPC. The condition $q \leq \min_{\l \in [L]} (n_\l-1)$ in Theorem~\ref{thm:NNPC} depends on the unknown quantities $n_\l$ and admits a large range of values for $q$ only if the clusters have balanced sizes, i.e., if $n_\l \approx \!N/L$, $\l \in [L]$. In practice, however, the performance of NNPC is observed to be quite robust w.r.t. the choice of $q$ even when the cluster sizes are imbalanced.

We would like to point out that virtually all existing analytical performance results for random process clustering apply to the asymptotic regime where $M \to \infty$, for $N$ fixed. The results that come closest to ours in spirit can be found in \cite{kakizawa1998discrimination, vilar2004discriminant}, where it is shown, in the asymptotic setting, that observations coming from different generative models can consistently (in the statistical sense) be discriminated via a PSD-based distance measure, provided that the PSDs of the generative models differ on a set of positive Lebesgue measure.

Finally, we note that the random process clustering problem considered here can be cast as a subspace clustering problem simply by interpreting the observations $\xh_i$ as vectors in $\reals^M$. Numerical results, not reported here, demonstrate, however, that NNPC clearly outperforms its subspace clustering cousin, the thresholding based subspace clustering (TSC) algorithm \cite{heckel2014robust}.
This is thanks to NNPC exploiting the stationarity of the generative models, a property that is usually inexistent in subspace clustering and is hence not taken into account by TSC.

\urldef{\code}\url{http://www.nari.ee.ethz.ch/commth/research/}
\vspace{-0.1cm}
\section[Numerical results]{Numerical results\footnote{\label{note1}Matlab code available at {\code}}}

\begin{figure}[h]
    \centering
    \tikzsetnextfilename{figsynthdata}
    \begin{tikzpicture}[scale=1] 
    
    \begin{axis}[name=plot1,
        	xlabel={\footnotesize $f$},
	xlabel absolute,
	xlabel style={yshift = 0.2cm},
    	width=0.5\columnwidth,
	height = 0.45\columnwidth,
	ymin = 0,
	ymax = 5,
	legend entries={ {$s_{\va_1,\vb_1}$}, {$s_{\va_2,\vb_2}$}, {$s_{\va_3,\vb_3}$}},
	xticklabel style={font=\footnotesize,},
	yticklabel style={font=\footnotesize,},
    	legend style={
                    font=\tiny,}
         ]
	{
		\addplot +[mark=none,solid,black] table[x index=0,y index=1]{psds.dat};
		\addplot +[mark=none,dashed,black] table[x index=0,y index=2]{psds.dat};
		\addplot +[mark=none,dotted,black] table[x index=0,y index=3]{psds.dat};
		}
	\end{axis}

       \begin{axis}[name=plot2,at={($(plot1.east)+(1cm,0)$)},anchor=west,
        	xlabel={\footnotesize $M$},
	xlabel absolute,
	xlabel style={yshift = 0.2cm},
    	width=0.65\columnwidth,
	height = 0.45\columnwidth,
	legend entries={ {NNPC, $\sigma^2 = 0$}, {KM, $\sigma^2 = 0$}, {NNPC, $\sigma^2 = 1/4$}, {KM, $\sigma^2 = 1/4$}}, 
	xticklabel style={font=\footnotesize,},
	yticklabel style={font=\footnotesize,},
    	legend style={
                    font=\tiny,}
         ]
	{
		\addplot +[mark=none,solid,black] table[x index=0,y index=2]{errs_sigmansq0.dat};
		\addplot +[mark=none,dashed,black] table[x index=0,y index=1]{errs_sigmansq0.dat};
		\addplot +[mark=none,dotted,black] table[x index=0,y index=2]{errs_sigmansq0.25.dat};
		\addplot +[mark=none,dashdotdotted,black] table[x index=0,y index=1]{errs_sigmansq0.25.dat};
		}
	\end{axis}
	
\end{tikzpicture}
\vspace{-0.35cm}
\caption{PSDs of the generative models (left) and average CE as a function of $M$ (right).} \label{fig:sim}
\vspace{-0.2cm}
\end{figure}
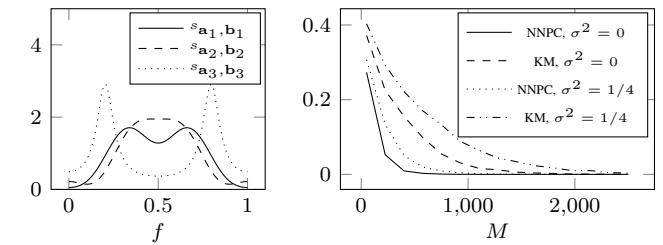

\paragraph*{Synthetic data} We evaluate the performance of NNPC and KM in terms of the clustering error (CE), i.e., the fraction of misclustered observations. We consider $L=3$ generative ARMA models with PSDs $s_{\va,\vb}(f) = \abss{\sum_{u = 1}^{m+1} b_u e^{-\i 2\pi u f}}^2 / \abss{\sum_{v = 1}^{n+1} a_v e^{-\i 2\pi v f}}^2$, where $\va$ and $\vb$ are coefficient vectors of length $n+1$ and $m+1$, respectively. We choose $\va_1 = 1, \vb_1 = [3/4\;1\;-7/4\;1/2], \va_2 = 1, \vb_2 =  [1/2\;5/4\;-3/2\;3/4]$, and $\va_3 = [1\;-1/5\;2/5\;1/10], \vb_3 = 1$, and then normalize the coefficient vectors to ensure that the processes have unit power. It can be seen in Figure~\ref{fig:sim} (left) that the process PSDs overlap significantly.
For the BT PSD estimator, we use a Gaussian window $g$ with standard deviation $50$ for both NNPC and KM. We sample $n_\l = 25$ independent observations from each generative model, and for each value of $M$ the CE is averaged over $200$ such problem instances. The number of generative models $L=3$ is assumed known. The performance of NNPC is rather insensitive to the choice of $q$ for $10 \leq q \leq 20$ (corresponding results are not shown here) and we set $q = 10$. Figure \ref{fig:sim} (right) shows the resulting average CE as a function of $M$. NNPC is seen to consistently outperform KM. For larger $M$ the performance difference between NNPC and KM becomes less pronounced. Note that for the choice of model parameters and window function in this experiment the clustering condition \eqref{eq:ClusCond} is not satisfied. 
\vspace{-0.3cm}
\paragraph*{Real data} We consider the problem of clustering sequences of human locomotion according to the activities performed. Specifically, we repeat the experiment from \cite{li2011time, khaleghi2012online}, which uses the Carnegie Mellon Motion Capture database\footnote{available at \url{http://mocap.cs.cmu.edu}} containing motion sequences of 149 subjects performing various activities. The motion sequences describe the positions of markers on different body parts over time, recorded using optical tracking. The experiment is based on subjects \#16 and \#35 for which the database contains 49 and 33 sequences, respectively, labeled either as ``walking'' or ``running''. 
We assume $L = 2$ to be known and we cluster the sequences describing the motion of the marker placed on the right foot of the subject. Differences in sequence lengths are accounted for by zero-padding to the maximum sequence length. Furthermore, we normalize the BT PSD estimates to unit power.
Table \ref{tab:CEMOCAP} lists the CE as well as the entropy of the clustering confusion matrix $S$ (defined in \cite[Sec. 6]{li2011time}) for $q=6$ (i.e., the value in $[\min_{\l \in [L]} (n_\l-1)]$ which yields the lowest CE and $S$ concurrently). Comparing $S$ to the corresponding values reported in \cite{li2011time, khaleghi2012online}, we find that for subject \#35 NNPC and KM perform better than the algorithm proposed in \cite{li2011time} and match the performance of the algorithm considered in \cite{khaleghi2012online}, while for subject  \#16 NNPC outperforms both of these algorithms as well as KM.
\vspace{-0.1cm}
\begin{table}[h]
\renewcommand{\arraystretch}{1.3}
\centering
\begin{tabular}{|c | c | c | c | c |}
\hline
& \multicolumn{2}{c |}{NNPC} & \multicolumn{2}{c|}{KM}\\
  subject & CE & $S$ & CE & $S$ \\
\hline
\#16 & 0.02 & 0.09 & 0.24 & 0.55 \\
\#35 & 0 & 0 & 0 & 0 \\
\hline
\end{tabular}
\vspace{0.2cm}
\caption{\label{tab:CEMOCAP} CE and $S$ for clustering human motion sequences}
\end{table}
\vspace{-0.6cm}

\appendix
\label{sec:ProofMain}
The central element in the proofs of Theorems \ref{thm:NNPC} and \ref{thm:KM} is the following result, proven at the end of the Appendix.
\begin{thm} \label{lem:DistCond}
Consider a data set $\cX$ generated according to the data model described in Section \ref{sec:mainres}. Then, the clustering condition \eqref{eq:ClusCond} implies
\begin{equation} \label{eq:DistRelation}
\min_{\substack{k, \l \in [L] \colon \\ k \neq \l}} \, \min_{\substack{i \in [n_\l], \\ j \in [n_k]}} d(\xkh_j,\xlh_i)
> \max_{\l \in [L]} \; \max_{\substack{ i,j \in [n_\l] \colon \\ i \neq j}} d(\xlh_i,\xlh_j)
\end{equation}
with probability at least $1 - 2N/M^2$.
\end{thm}
Theorem \ref{lem:DistCond} essentially says that under the clustering condition \eqref{eq:ClusCond} observations stemming from the same generative model are closer (in terms of the distance measure $d$) than observations stemming from different generative models. We now show how Theorems \ref{thm:NNPC} and \ref{thm:KM} follow from Theorem \ref{lem:DistCond}.
\paragraph*{Proof of Theorem \ref{thm:NNPC}} \label{sec:ThmNNPCproof}
The NFC property for $q \leq \min_{\l \in [L]} (n_\l - 1)$ is a direct consequence of \eqref{eq:DistRelation}, which by Theorem \ref{lem:DistCond}, is implied by the clustering condition \eqref{eq:ClusCond}.

\paragraph*{Proof of Theorem \ref{thm:KM}}
\label{sec:ThmKMproof}
The proof is effected by showing that in Step 3 KM selects an observation with a different generative model in every iteration, i.e., $\{ \xh_{c_\l} \}_{\l = 1}^L$ contains exactly one observation of each generative model, provided that the clustering condition \eqref{eq:ClusCond} and hence, by Theorem \ref{lem:DistCond}, \eqref{eq:DistRelation} holds. The argument is then concluded by noting that, again by \eqref{eq:DistRelation} and hence by \eqref{eq:ClusCond}, the partition $\hat \cX_1,\dots,\hat \cX_L$ obtained in Step 4 corresponds to the true partition $\cX_1,\dots,\cX_L$.

Suppose that after the $p$th iteration in Step 3 of KM the observations $\xh_{c_\l}, \l \in [p],$ all come from different generative models, and assume w.l.o.g. that the generative model underlying $\xh_{c_\l}$ has index $\l$, for $\l \in [p]$. For iteration $p+1$, it follows from \eqref{eq:DistRelation} that
\begin{align}
&\max_{i \in [N]} \min_{\l \in [p]} d(\xh_i, \xh_{c_\l}) \nonumber \\
& = \! \max \! \Bigg\{ \! \max_{\substack{k \in [p], \\ i \in [n_k]}} \, \min_{\l \in [p]} d(\xkh_i\!,\xlh_{c_\l}), \!
\max_{\substack{k \in [L] \setminus[p], \\ i \in [n_k]}} \, \min_{\l \in [p]} d(\xkh_i\!,\xlh_{c_\l}) \Bigg\} \nonumber \\ 
&\, = \max \Bigg\{ \! \! \! \! \! \! \underbrace{\max_{\substack{k \in [p],\\ i \in [n_k]}} d(\xkh_i,\xkh_{c_k})}_{\leq \underset{\l \in [L]}{\max} \; \underset{\substack{i,j \in [n_\l] \colon \\ i \neq j}}{\max} d(\xlh_i,\xlh_j)}\! , 
\underbrace{ \max_{\substack{k \in [L] \setminus[p], \\ i \in [n_k]}} \; \min_{ \l \in [p]} d(\xkh_i,\xlh_{c_\l})}_{ \geq \underset{\substack{k, \l \in [L] \colon \\ k \neq \l}}{\min} \; \underset{\substack{i \in [n_\l], \\ j \in [n_k]}}{\min} d(\xkh_i,\xlh_j)} \Bigg\} \nonumber \\
&\, = \max_{\substack{k \in [L] \setminus[p],\\ i \in [n_k]}} \, \min_{\l \in [p]} d(\xkh_i,\xlh_{c_\l}). \label{eq:MinMaxLast}
\end{align}
Note that in the maximization in \eqref{eq:MinMaxLast} $k$ runs over $[L] \setminus [p]$ which means that $\xh_{c_{p+1}}$ is guaranteed to have a generative model that is different from those underlying $\xh_{c_1}, \dots, \xh_{c_p}$. Iterating the preceding argument for $p = 2, \dots, L$, we find that, indeed, $\{ \xh_{c_\l} \}_{\l = 1}^L$ contains exactly one observation of each generative model.

\paragraph*{Proof of Theorem \ref{lem:DistCond}}
Let $\slt(f) \defeq \sl(f) + \wl(f)$ with $\wl(f) = \sigma^2$, $f \in [0,1)$, for all $\l \in [L]$, be the PSD of $\Xlt$ and $\rlt$ the corresponding ACF. Define $\el_i(f) \defeq \slh_i (f) - \slt(f)$ and set $\varepsilon \defeq \max_{\l \in [L], i \in [n_\l]} \sup_{f \in [0,1)} \abss{\el_i(f)}$. We have for all $k, \l \in [L], i \in [n_\l], j \in [n_k]$, that
\begin{align}
d(\xkh_j, \xlh_i) &= \frac{1}{2} \int_0^1 \abs{\skh_j(f) - \slh_i(f)} \d f \nonumber \\
&= \frac{1}{2} \int_0^1 \left\vert \sk(f) + \wk(f) + \ek_j(f) \right. \nonumber \\ & \qquad \qquad \quad \left. - (\sl(f) + \wl(f) + \el_i(f) ) \right\vert \d f \nonumber \\
&\leq d(\Xk,\Xl) + \frac{1}{2} \int_0^1 \abs{\ek_j(f)}  \d f \nonumber \\ &\quad + \frac{1}{2} \int_0^1 \abs{\el_i(f)} \d f  \label{eq:UBTriang1}\\
&\leq d(\Xk,\Xl) + \varepsilon. \label{eq:DistUB}
\end{align}
Reversing the triangle inequality leading to \eqref{eq:UBTriang1} it follows similarly that
\begin{equation}
d(\xkh_j, \xlh_i) \geq d(\Xk,\Xl) - \varepsilon \label{eq:DistLB}
\end{equation}
for all $k, \l \in [L], i \in [n_\l], j \in [n_k]$. 
Replacing $d(\xlh_i,\xlh_j)$ on the RHS of \eqref{eq:DistRelation} by the upper bound in \eqref{eq:DistUB} and using the lower bound in \eqref{eq:DistLB} on the LHS of \eqref{eq:DistRelation}, we find that \eqref{eq:DistRelation} is implied by
\begin{equation} \label{eq:DistCondLBeps}
\min_{k, \l \in [L] \colon k \neq \l} d(\Xk, \Xl) > 2 \varepsilon.
\end{equation}

We continue by upper-bounding $\varepsilon$. To this end, define $\mQ_m \in \{0,1\}^{M \times M}$ as $(\mQ_m)_{u,v} = 1$, if $v - u = m$, and $(\mQ_m)_{u,v} = 0$, otherwise, $\mc M \defeq \{ -M +1, -M + 2, \dots, M - 1\}$, and
$\mG(f) \defeq \sum_{m \in \mc M} g[m] \cos(2 \pi f m) \mQ_m$, 
i.e., $\mG_{u,v}(f) = \mG_{v,u}(f) = g[v-u] \cos(2 \pi f (v-u))$. Now, with $\vx \in \reals^M$ the random vector containing the elements of $\xlh_i$, it holds for $m \in \mc M$ that
\begin{align}
\rlh_i [m] = \frac{\transp{\vx} \mQ_m \vx}{M} \qquad \text{and} \qquad \rlt [m] = \frac{\E{\transp{\vx} \mQ_m \vx}}{M-\abs{m}}. \nonumber
\end{align}
With these relations we have
\begin{align}
&\sup_{f \in [0,1)} \abs{\el_i(f)} \nonumber = \sup_{f \in [0,1)} \abs{\slh_i(f) - \slt(f)} \nonumber \\
&= \sup_{f \in [0,1)} \bigg\vert \sum_{m \in \mc M} g[m] \rlh_i [m] e^{-\i 2 \pi f m} -  \sum_{m \in \mb Z} \rlt [m] e^{-\i 2 \pi f m} \bigg\vert \nonumber\\
&= \sup_{f \in [0,1)} \bigg\vert\underbrace{\sum_{m \in \mc M } \frac{g[m]}{M}\left( \transp{\vx} \mQ_m \vx - \E{ \transp{\vx} \mQ_m \vx } \right) e^{-\i 2 \pi f m}}
_{\substack{\frac{1}{M} \big( \transp{\vx} \left( \sum_{m \in \mc M} g[m] \cos(2 \pi fm) \mQ_m \right) \vx \\ \hspace{2cm} - \mb E \left[ \transp{\vx} \left( \sum_{m \in \mc M} g[m] \cos(2 \pi fm) \mQ_m \right) \vx \right] \big)}} \nonumber \\ 
& \quad + \sum_{m \in \mc M} \underbrace{\frac{g[m]}{M} \E{ \transp{\vx} \mQ_m \vx }}_{= g[m] \left( 1 - \frac{\abs{m}}{M} \right) \rlt [m]} \! \! e^{-\i 2 \pi f m} - \sum_{m \in \mb Z} \rlt [m] e^{-\i 2 \pi f m} \bigg\vert \label{eq:SupELIQuadForm} \\[-0.3cm]
&= \sup_{f \in [0,1)} \bigg\vert \underbrace{ \frac{1}{M} \left( \transp{\vx} \mG(f) \vx - \E{ \transp{\vx} \mG(f) \vx} \right)}_{ \eqdef \al_i(f)}  \nonumber \\ 
& \hspace{3.5cm} - \sum_{m \in \mb Z} h[m] \rlt [m] e^{-\i 2 \pi f m} \bigg\vert \label{eq:SupELIQuadFormTot}  \\
&\leq \sup_{f \in [0,1)} \left\vert \al_i(f) \right\vert + \sum_{m \in \mb Z} \big\vert  h[m] \big\vert \big\vert \rl[m] \big\vert + \underbrace{\big\vert h[0] \big\vert}_{=0} \sigma^2 \nonumber \\
&= \sup_{f \in [0,1)} \abs{\al_i(f)} + \mu_{\max}, \label{eq:SupELIFinal}
\end{align}
where we used the fact that $g[m] (\transp{\vx} \mQ_m \vx - \E{ \transp{\vx} \mQ_m \vx })$ is a real-valued even sequence, employed the definition of $h[m]$ in \eqref{eq:weightedwindow} in the step leading from \eqref{eq:SupELIQuadForm} to \eqref{eq:SupELIQuadFormTot}, and invoked $g[0] = 1$ (i.e., $h[0]=0$) as well as $\rlt[m] = \rl[m] + \sigma^2 \delta[m]$ to obtain the last inequality.
It follows from \eqref{eq:SupELIFinal} that $\varepsilon \leq \max_{\l \in [L], i \in [n_\l]} \sup_{f \in [0,1)} \big\vert \al_i (f) \big\vert + \mu_{\max}$ and hence \eqref{eq:ClusCond} implies \eqref{eq:DistRelation} via \eqref{eq:DistCondLBeps} on the event
\begin{equation*}
	\mc F \! \defeq \! \left\{ \! \max_{\l \in [L], i \in [n_\l]} \sup_{f \in [0,1)} \! \abs{\al_i (f)} \! < \! 4 A (B+ \sigma^2) \sqrt{\frac{2 \log M}{M}} \right\} \!.
\end{equation*}
It remains to lower-bound $\PR{\mc F}$, which will be accomplished by upper-bounding the tail probability of the random variables $\al_i (f)$. For fixed $f$, the distribution of $\al_i (f)$ obeys
\begin{align}
\al_i (f) \sim \frac{1}{M} \left( \transp{\vy} \transp{\mC} \mG(f) \mC \vy - \E{ \transp{\vy} \transp{\mC} \mG(f) \mC \vy } \right), \nonumber
\end{align}
where the entries of $\vy$ are independent standard normal random variables and $\mC = (\mR + \sigma^2 \mI)^{1/2} \in \reals^{M \times M}$ with $\mR_{u,v} = \rl[v-u]$ the (Toeplitz) covariance matrix corresponding to $M$ consecutive elements of $\Xl$. Setting $\mB \defeq \transp{\mC} \mG(f) \mC$, we can establish a bound on the tail probability of $\al_i$ by invoking a well-known concentration inequality for quadratic forms in Gaussian random vectors \cite[Lem. 1]{demanet2012matrix}, namely
\begin{align}
&\mathrm{P} \Big[ \abs{ \transp{\vy} \mB \vy - \E{\transp{\vy} \mB \vy}} \nonumber \\
& \hspace{1.5cm} \geq \norm[F]{\mB + \transp{\mB}}\sqrt{\delta} + 2 \norm[2 \to 2]{\mB} \delta \Big] \leq 2 e^{-\delta}. \label{eq:GaussQuadForm}
\end{align}
Next, we note that $\norm[F]{\smash{\mB + \transp{\mB}}} \leq 2 \norm[F]{\mB} \leq 2 \sqrt{M} \norm[2 \to 2]{\mB}$ and $\norm[2 \to 2]{\mB} \leq \norm[2 \to 2]{\mC}^2 \norm[2 \to 2]{\mG(f)} = \norm[2 \to 2]{\smash{\mR + \sigma^2 \mI}} \norm[2 \to 2]{\mG(f)} \leq A (B + \sigma^2)$, where we used that both $\mR$ and $\mG(f)$ are symmetric Toeplitz matrices and hence, by \cite[Lem. 4.1]{gray2006toeplitz}, $\norm[2 \to 2]{\mR} \leq \sup_{f \in [0,1)} \sl(f) \leq B$ and $\norm[2 \to 2]{\mG(f)} \leq \sup_{f' \in [0,1)} g(f+f') = \sup_{f' \in [0,1)} g(f') \leq A$ (here, the frequency shift by $f$ due to the $\cos$-factors in the elements of $\mG(f)$ does not affect the supremum because $g(f)$ is $1$-periodic). Now, setting $\delta = 2 \log (M)$ and using $\delta/M < \sqrt{\delta/M} < 1$, for $M \geq 1$, \eqref{eq:GaussQuadForm} yields
\begin{equation}
\PR{\sup_{f \in [0,1)} \abs{\al_i (f)} \geq 4 A (B+ \sigma^2) \sqrt{\frac{2 \log M}{M}}} \leq \frac{2}{M^2}. \nonumber
\end{equation}

Finally, it follows from a union bound argument that
\begin{align}
\PR{\mc F} \! & \geq \! 1 - \! \! \sum_{\substack{\l \in [L], \\ i \in [n_\l]}} \! \! \PR{\sup_{f \in [0,1)} \! \abs{\al_i (f)} \! \geq \! 4 A (B+ \sigma^2) \sqrt{\frac{2 \log M}{M}}}  \nonumber \\ 
& > \! 1 - \frac{2 N}{M^2}. \nonumber
\end{align}

\bibliographystyle{IEEEtran}
\bibliography{IEEEabrv,processclustering.bib}
%
%
%

\end{document}